# Social Support Detection from Social Media Texts

Zahra Ahani[a,∗], Moein Shahiki Tash[a,1,∗], Fazlourrahman Balouchzahi[a], Luis Ramos[a], Grigori Sidorov[a], Alexander Gelbukh[a]

[a]*Instituto Politécnico Nacional (IPN), Centro de Investigación en Computación (CIC), Mexico City, Mexico,*

**Abstract**

Social support, conveyed through a multitude of interactions and platforms such as social media, plays a pivotal role in fostering a sense of belonging, aiding resilience in the face of challenges, and enhancing overall well-being. This paper introduces Social Support Detection (SSD) as a Natural language processing (NLP) task aimed at identifying supportive interactions within online communities. The study presents the task of Social Support Detection (SSD) in three subtasks: two binary classification tasks and one multiclass task, with labels detailed in the dataset section. We conducted experiments on a dataset comprising 10,000 YouTube comments. Traditional machine learning models were employed, utilizing various feature combinations that encompass linguistic, psycholinguistic, emotional, and sentiment information. Additionally, we experimented with neural network-based models using various word embeddings to enhance the performance of our models across these subtasks.The results reveal a prevalence of group-oriented support in online dialogues, reflecting broader societal patterns. The findings demonstrate the effectiveness of integrating psycholinguistic, emotional, and sentiment features with n-grams in detecting social support and distinguishing whether it is directed toward an individual or a group. The best results for different subtasks across all experiments range from 0.72 to 0.82.

## 1. Introduction

Social support is the provision of behaviors, communication, and interactions that convey care and value to individuals, fostering a sense of belonging and aiding in coping with life's challenges (Ko et al., 2013). Social support manifests in diverse ways, ranging from expressions of care and encouragement to practical assistance or guidance. Recognizing the presence of supportive individuals who offer various forms of aid can serve as a buffer against stress and safeguard both emotional and physical health. The support patients receive from shared content plays a crucial role in enhancing their self-care practices and overall health results (Jadad et al., 2000). Specifically, individuals coping with chronic illnesses, disabilities, or cancer often find social media platforms invaluable, as they offer opportunities to connect with peers or professionals for guidance in managing their long-term conditions effectively (Merolli et al., 2013).

In recent years, a heightened awareness of the detrimental impacts of hate speech, abusive language, and misogyny on social media platforms has led to a surge in research efforts focused on their detection through Natural Language Processing (NLP) (Shashirekha, 2020). While social media platforms offer users the freedom and anonymity to express their opinions and engage in instant feedback, this liberty also fosters an environment where individuals may exploit the platform to propagate discriminatory or harmful views targeting specific demographics (Chakravarthi, 2020). Consequently, developing tools and techniques for detecting and mitigating such content has become imperative in creating safer digital environments and promoting respectful online discourse.

However, some argue that this approach can infringe on users' freedom of expression (Chakravarthi, 2020; Balouchzahi et al., 2023). Instead of solely focusing on identifying and removing negative content, an alternative strategy could involve promoting positive interactions and supporting content that contributes to social good. By encouraging and amplifying constructive and respectful communication, social media platforms can foster a more positive online environment while still respecting users' rights to freely express their opinions. This dual approach not only mitigates the spread of harmful content but also actively contributes to a more supportive and inclusive digital community.

Despite the importance of promoting positive and supportive content, not many tasks have been done in this research area. In response to these challenges, our proposed approach offers an alternative but under-explored strategy to combat the negative atmosphere on social media platforms by promoting social support comments. Rather than solely focusing on identifying and filtering out negative content, our approach seeks to cultivate a more positive and supportive online environment by encouraging users to provide emotional comfort, encouragement, and advice to those facing challenges.

Online social support encompasses the assistance and emotional comfort offered via digital platforms such as social media, forums, and messaging apps. This type of support is crucial for individuals and groups facing various challenges, such as victims of wars, black community, or minorities. Through these digital channels, individuals can connect with others who

---

share similar experiences, access valuable resources, and receive empathy and encouragement. The anonymity and accessibility of online support networks often make them a vital lifeline for those who might not have access to traditional forms of support. Additionally, these platforms can provide real-time assistance, foster a sense of community, and help reduce feelings of isolation and loneliness. A detailed definition and social support are presented in section 2.

By fostering a culture of support and empathy, we aim to alleviate stress, enhance coping mechanisms, and promote overall well-being in the face of adversity, ultimately creating a more inclusive and nurturing online community.

Inspired by tasks in hate speech detection (Madhu et al., 2023b,a), an opposite task, social support detection (SSD) from text, is proposed, which is modeled into a three-step classification task. Data was collected from YouTube, starting with the viewing of numerous videos where comments with supportive subjects were noted. Subsequently, comments were randomly checked and collected, and after preprocessing, a random sample of 10,000 comments was selected for manual annotation. This resulted in 2,236 supportive comments and 7,762 non-supportive comments. Following data preparation, experiments were conducted using traditional machine learning models, including Logistic Regression (LR), Support Vector Machine with radial basis function kernel (SVM(rbf)), Support Vector Machine with the linear kernel (SVM(linear)), Decision Trees (DT), and Random Forest Classifier (RFC). Three different feature sets were used: LIWC, Unigrams vectorized with TF-IDF, and LIWC+unigrams. We also used different word embeddings (GloVe and FastText) and model architectures (CNN and BiLSTM) to make predictions and compare them with other models.

The main contributions of this study are listed below:

- Study of social support for social good as a novel task in NLP,

- Developing annotations guidelines and generating the first specific social support detection dataset in English,

- Study of psycholinguistic features of social support for different levels: group, individual, and target groups,

- Providing benchmark experiments using traditional machine learning models and linguistic and psycholinguistic features.

## 2. Definitions

Albrecht and Goldsmith (2003) define social support as verbal and nonverbal communication between recipients and providers that reduces uncertainty about the situation, the self, the other, or the relationship and functions to enhance the perception of personal control in one's experience.

Although social support is helpful during stressful situations, Barnes and Duck (1994) pointed out that the exchange of support does not only manifest during the crisis but is also an everyday occurrence in personal relationships.

Cutrona and Suhr (1992) delineate a social support framework comprising five primary categories: informational, emotional, esteem, social network, and tangible support. Informational support entails messages conveying knowledge or advice, emotional support involves expressions of care and empathy, esteem support boosts self-worth and abilities, social network support fosters a sense of belonging within a group, and tangible support entails physically providing goods or services as needed.

The current research aligns with (Barnes and Duck, 1994) definition of social support and views the exchange of comments and feedback between users and audiences as a form of social support occurring within their communication. Social support refers to "information leading the subject to believe that he is cared for and loved, esteemed, and a member of a network of mutual obligations" (Cobb, 1976). It is formed by the exchange of resources (i.e., verbal and nonverbal messages) between two or more individuals Shumaker and Brownell (1984).

Studies have demonstrated that social support offers advantages to patients, such as dealing with challenging life circumstances (Thoits, 1982), enhancing compliance with recommended treatment plans McCorkle et al. (2008), and fostering better mental health Cohen and Wills (1985). In the management of chronic illnesses, social support plays a critical role in encouraging healthy behaviors and attaining favorable health results for patients. For instance, McCorkle et al. (2008). discovered that social support enhances individuals' quality of life and diminishes psychological distress among those facing severe mental health issues.

Hence, social support is defined as "the emotional, informational, or practical assistance offered by others, including peers or community members. This aid can be extended to individuals or groups, such as women, religious communities, or racial minorities like black community, aiding them in navigating challenges, enhancing their overall well-being, and fostering resilience."

## 3. Related work

Despite the critical importance of promoting positive and supportive content, research in this area remains relatively sparse. This concept serves as a counterpoint to hate speech. While there is no directly comparable work specifically focused on positive content within NLP, related research has been conducted on tasks such as hope speechArif et al. (2024), which offers some insights into this area. Therefore, in this section, we summarize some of the hate and hope speech research.

The phenomenon of hate speech and violent communication online is commonly referred to as cyberhate (Miró-Llinares and Rodríguez-Sala, 2016). It involves the use of electronic communication technologies to propagate discriminatory or extremist messages, targeting not only individuals but also entire communities Blaya (2019). Hate speech encompasses various linguistic styles and actions, including insults, provocation, and aggression (Anis, 2017). It can be categorized into different types, such as gendered hate speech, which targets specific genders or promotes misogyny, religious hate speech, which discriminates against various religious groups, and racist hate speech, which involves racial discrimination and prejudice against particular ethnicities or regions (Chetty and Alathur, 2018).

The feasibility of utilizing domain-specific word embeddings as features and a bidirectional LSTM-based deep model as a classifier for the automatic detection of hate speech was studied by (Saleh et al., 2023). Three datasets were used, with a total collection comprising 21,514 non-hate and 27,085 hate instances. To ensure balanced data, 16,260 instances for each label were used. This approach facilitated the detection of coded language by appropriately ascribing negative connotations to words. Additionally, the applicability of the transfer learning language model (BERT) to the hate speech classification task was investigated, given its high-performance results across various NLP tasks. Experimental findings indicated that the combination of domain-specific word embeddings with the bidirectional LSTM-based deep model achieved an F1 score of 93%, while BERT achieved an F1 score of 96% when applied to a combined balanced dataset sourced from existing hate speech datasets.

Hasan et al. (2022) proposed a deep learning model that was utilized for the classification of sentimentsTash et al. (2024c) in two distinct analyses. In the first analysis, tweets were categorized based on hate speech targeting migrants and women. In the second analysis, detection is carried out using a deep learning modelAhani et al. (2024a) to ascertain whether the hate speech is perpetrated by a single user or a group of users. The dataset is divided into two segments: a training dataset comprising 31,963 records and a testing dataset containing 17,198 records. These datasets are further partitioned into English and Spanish datasets. The training dataset is employed for model development, while the testing dataset is utilized for model validation. Word embedding was implemented during text analysis, employing a combination of deep learning models, including BiLSTM, CNN, and MLP. These models are integrated with word embedding techniques such as inverse glove (global vector), document frequency (TF-IDF), and transformer-based embedding. A comparison of models based on the English dataset for hate speech classification against women is presented in Table 1.

Albadi et al. (2018) details the creation of the first publicly available Arabic dataset annotated for religious hate speech detection and the development of the initial Arabic lexicon containing terms commonly encountered in religious discussions, along with their polarity and strength scores. Various classification models were then constructed utilizing lexicon-based, n-gram-based, and deep-learning-based methodologies. A comprehensive comparison of model performances on a completely novel, unseen dataset is subsequently presented. It is observed that a straightforward Recurrent Neural Network (RNN) architecture with Gated Recurrent Units (GRU) and pre-trained word embeddings can effectively identify religious hate speech, achieving an AUROC of 0.84. In November 2017, a collection of 6,000 Arabic tweets, consisting of 1,000 tweets for each of the six religious groups, was obtained using Twitter's search API. This paper specifically concentrates on the four predominant religious beliefs in the Middle East, namely Islam (93.0%), Christianity (3.7%), Judaism (1.6%), and Atheism (0.6%). Given that Islam is the most prevalent religion in the region, both Sunni and Shia, the two main sects of Islam accounting for 87-90% and 10-13% of all Muslims, respectively, are included. The best results were related to the GRU-based RNN, which achieved a macro F1 score of 0.77.

The specificities of online hate speech against Afro-descendant, Roma, and LGBTQ+ communities in Portugal are addressed in (Carvalho et al., 2023). The analysis of CO-HATE, a corpus comprising 20,590 YouTube comments manually annotated following detailed guidelines, forms the basis of the research. Significant differences in hate speech prevalence across these communities were revealed, with a higher incidence targeting the Roma community. Despite efforts to model covert hate speech, its complexity poses challenges for NLP systems. The importance of investigating the discursive and rhetorical strategies underlying covert hate speech is highlighted. Covert hate speech, which often employs rhetorical devices such as irony and sarcasm, aims to discriminate against targets and invokes emotions through appeals to extreme right-wing populist ideology. The inclusion of annotators from targeted groups emphasizes the multiplicity of perspectives, which is crucial for understanding hate speech dynamics. Inclusive approaches to improve hate speech detection models, considering diverse perspectives, are suggested by the study.

Chakravarthi (2020) Hope Speech for Equality, Diversity, and Inclusion (HopeEDI) dataset by collecting user comments from YouTube. These comments were manually categorized as containing hopeful content or not, resulting in 28,451 comments in English, 20,198 in Tamil, and 10,705 in Malayalam. Comments in languages other than the intended ones were labeled as "other languages." Table 2 presents detailed statistics of this dataset. Various machine learning algorithmsRamos et al. (2024a,b) such as Support Vector Machine (SVM), Multinomial Naive Bayes (MNB), K-Nearest Neighbors (KNN), Decision Tree (DT), and Logistic Regression (LR) were employed. The results indicate that DT outperformed other models for English and Malayalam, achieving average weighted F1 scores of 0.46 and 0.56, respectively. However, for Tamil texts, LR achieved the highest average weighted F1 score of 0.55.

Balouchzahi et al. (2023) introduced PolyHope, the first multiclass hope speech detection dataset in English. The dataset creation process involved the collection of approximately 100,000 English tweets, which were preprocessed to yield around 23,000 tweets. A random subset of 10,000 tweets was subsequently selected for annotation, resulting in final statistics of Hope=4175 and Not-Hope=4081 post-annotation. They further fine-grained the type of hope into General, Realistic, and Unrealistic hopes. To assess the dataset's performance, various baseline models were evaluated using diverse learning approaches, including traditional machine learningZamir et al. (2024b), deep learningZamir et al. (2024a); Ahani et al. (2024c), and transformer-based methods. The top-performing models for each learning approach demonstrated the average macro F1 scores for both binary and multiclass classification tasks on the PolyHope dataset, with transformers achieving bet-ter results, scoring 0.85 for binary classification and 0.72 for multiclass classification.

| Language | Model | Task | Macro F1 score |
|---|---|---|---|
| **English** | Transformer-CNN and MLP | hate speech classification against woman | 93.4 |
| | | hate speech detection by the individual or group of people | 94.19 |
| **Spanish** | | hate speech classification against woman | 93.52 |
| | | hate speech detection by the individual or group of people | 93.82 |

Table 1: Statistics and Data Sources

| Class | English | Tamil | Malayalam |
|---|---|---|---|
| **Hope** | 2,484 | 7,899 | 2,052 |
| **Not Hope** | 25,940 | 9,816 | 7,765 |
| **Other languages** | 27 | 2,483 | 888 |
| **Total** | 28,451 | 20,198 | 10,705 |

Table 2: HopeEDI dataset statistics

This paper Palakodety et al. (2019) delves into an unfolding international crisis through a substantial corpus derived from comments on YouTube videos, consisting of 921,235 English comments contributed by 392,460 users out of a total of 2.04 million comments posted by 791,289 users across 2,890 videos. Three primary contributions are highlighted. Firstly, the effectiveness of polyglot word embeddings in revealing precise language clusters is emphasized, leading to the development of a document language identification technique requiring minimal annotation. Its applicability and usefulness across various datasets involving multiple low-resource languages are showcased. Secondly, temporal trends in pro-peace and pro-war sentiment are examined, noting that during periods of heightened tension between the two nations, pro-peace sentiment in the corpus reached its peak. Lastly, amidst politically charged discussions in a volatile scenario on the brink of potential conflict, the significance of automatically identifying user-generated web content capable of diffusing hostility is emphasized. The task of hope-speech detection is introduced, and the best performance achieved, with an F1 score of 78.51% ± 2.24%, using n-grams is reported.

## 4. Dataset development

### 4.1. Data collection and processing

This research focuses on analyzing data collected from YouTube comments across 17 videos spanning various categories such as nation, black community, women, religion, LGBTQ, and others. The video selection was based on topics potentially related to supportive content. For instance, videos concerning the war between Israel and Palestine, events involving the Black Community, Christiano Ronaldo, LGBTQ+ issues, and Women were chosen. Initially, we amassed 66,272 comments. After filtering out duplicate and non-English comments, the dataset was refined to 42,695 comments.

Subsequently, we proceeded to select 5,000 comments containing specific keywords, and another 5,000 comments were chosen randomly The following keywords were selected for analysis: "support", "stay strong", "I'm here to help", "I believe in you", and "inspiring", and their synonyms. These keywords capture the essence of social support and convey various aspects of emotional assistance.

It is worth noting that comments associated with the selected videos underwent no additional filtering or selection process. This approach enabled us to accurately gauge the distribution of supportive comments for each video while also exploring related aspects.

### 4.2. Annotator selection

For annotator selection, two annotators, one male and one female, both holding master's degrees in computer science and possessing proficient English language skills, were hired. Initially, each annotator was provided with 100 sample tweets and detailed annotation guidelines. Subsequently, the labeled samples from two annotators were analyzed. Individual meetings and interviews were then conducted with these annotators to address any confusion and ensure a thorough understanding of the annotation task.

As a third annotator, one of the authors of the paper, who had comprehensive knowledge of the concept, was selected. This annotator, a Ph.D. student in natural language processing with advanced English proficiency, contributed to the annotation process.

Finally, the 10,000 data points were divided into five parts, with each annotator completing one part. After each part, comments were randomly selected to verify the accuracy of their annotations.

### 4.3. Annotation guidelines

The SSD task was structured as a three-step classification process. First, supportive comments were identified. Next, it was determined whether these supportive comments were directed toward an individual, a group, or a community. Finally, if the supportive comment was identified as being directed toward a group, the specific group was further identified. The guidelines for this process are described below.

- **Subtask 1 - Binary social support detection:** In this subtask, a given text is classified as either supportive or non-supportive:

  – **Social Support (label = SS):** Supportive statements promote understanding, empathy, and positive action. Therefore, a supportive comment is a statement or message that offers support, encouragement, admiration, or assistance to individuals or groups that are encountering difficulties or have accomplished something noteworthy. These comments aim to provide emotional support, boost morale, or acknowledge the achievements of others.

- **Not Social Support (label = NSS):** The text does not convey any form of support, admiration, or encouragement.

• **Subtask 2 - Individual vs. Group:** In this subtask, each pre-identified in Subtask 1 supportive comment is further classified as support for an individual or support for a group:

- **Individual:** If the text expresses support for a specific person or individual (e.g., Cristiano Ronaldo, Trump), it is labeled as Support for Individual.
- **Group:** If the text expresses support for a group of people, community, tribe, nation, etc. (e.g., Muslims, Real Madrid, Black nations, LGBTQ), it is labeled as Support for Group.

• **Subtask 3 - Multiclass SS for Groups:** In this subtask, we aim to identify which community or group of people are targeted for social support by classifying the group supportive comments identified in Subtask 2 into the following categories:

- **Women:** The text expresses support for women and promotes women's rights and feminism.
- **Black community:** The text expresses support for the black community and promotes black community rights.
- **LGBTQ:** The text expresses support for the LGBTQ community and promotes LGBTQ community rights.
- **Religion:** The text expresses support for a religion and its rights.
- **Other:** The text expresses support for a community other than those listed above.

*4.4. Annotation procedure*

Detailed annotation guidelines and sample data were provided to the three chosen annotators to facilitate the creation of the proposed dataset. The annotators followed a structured process, as illustrated in Figure 1. Initially, they determined whether comments expressed support, concern, or care. If affirmative, they proceeded to a second level of analysis, distinguishing whether the support was directed towards an individual or a group. In cases where it pertained to a group, annotators further specified the group's affiliation, such as Nation, Religion, Black Community, Women, LGBTQ, or Other. Conversely, if the comment did not exhibit support, annotators labeled it as Non-supportive. This meticulous process ensured comprehensive annotation and dataset quality.

*4.5. Inter-annotator agreement*

Inter-annotator agreement (IAA) assesses how much annotators agree, factoring in chance agreement. Cohen's Kappa Coefficient scores of 85% for subtask 1, 82% for subtask 2, and 65% for subtask 3 demonstrate the robustness of the datasets, reflecting the rigorous annotation process.

*4.6. Statistics of the dataset*

Table 3 showcases the SSD dataset, which is divided into three tasks with varying sample sizes. In Task 1, the number of Non-Social Support samples (7,762) greatly exceeds Social Support samples (2,236), indicating a higher prevalence of non-supportive comments. Task 2 shows more comments related to groups (1,805) than individuals (417), reflecting a tendency for discussions to focus on collective rather than individual subjects. Task 3 reveals a wide distribution across specific categories, with "Nation" having the most samples (980) and "Religion" the fewest (18). The higher number of comments about "Nation" can be attributed to the trending topic at the time of data collection, specifically the Israel-Palestine conflict. "Other" also has a significant number of samples (512) because it encompasses various additional categories within supportive comments. These differences likely stem from the inherent interest and sensitivity of topics, the nature of the videos, and the sampling methods used. You can find the statistics of the dataset in Table 3.

| Tasks | Category | Number of samples |
|---|---|---|
| Subtask1 | Social Support | 2236 |
| | Not Social Support | 7762 |
| Subtask2 | Individual | 417 |
| | Group | 1805 |
| Subtask3 | Nation | 980 |
| | Other | 512 |
| | LGBTQ | 155 |
| | Black Community | 115 |
| | Women | 24 |
| | Religion | 18 |

Table 3: Statistics of SSD dataset

**5. Feature Analysis**

This section delves into the complex interplay of psycholinguistic, emotionalTash et al. (2024b), and sentiment features utilized in our research. It not only outlines these features but also explores how they relate to various forms of social support. By understanding these dynamics, we gain a nuanced understanding of how language and emotions intersect with mechanisms of social support. Psycholinguistic attributes involve linguistic cues intertwined with psychological processes, extracted using LIWC software from social supportive comments. Emotional features encompass the expression of emotions Tash et al. (2024d), intensity, and valence within communication, shedding light on the emotional dynamics of supportive interactions. Additionally, sentiment features reveal the overall sentiment conveyed in text, whether positive, negative, or neutral, offering insights into the prevailing tone and attitude in supportive discourse.

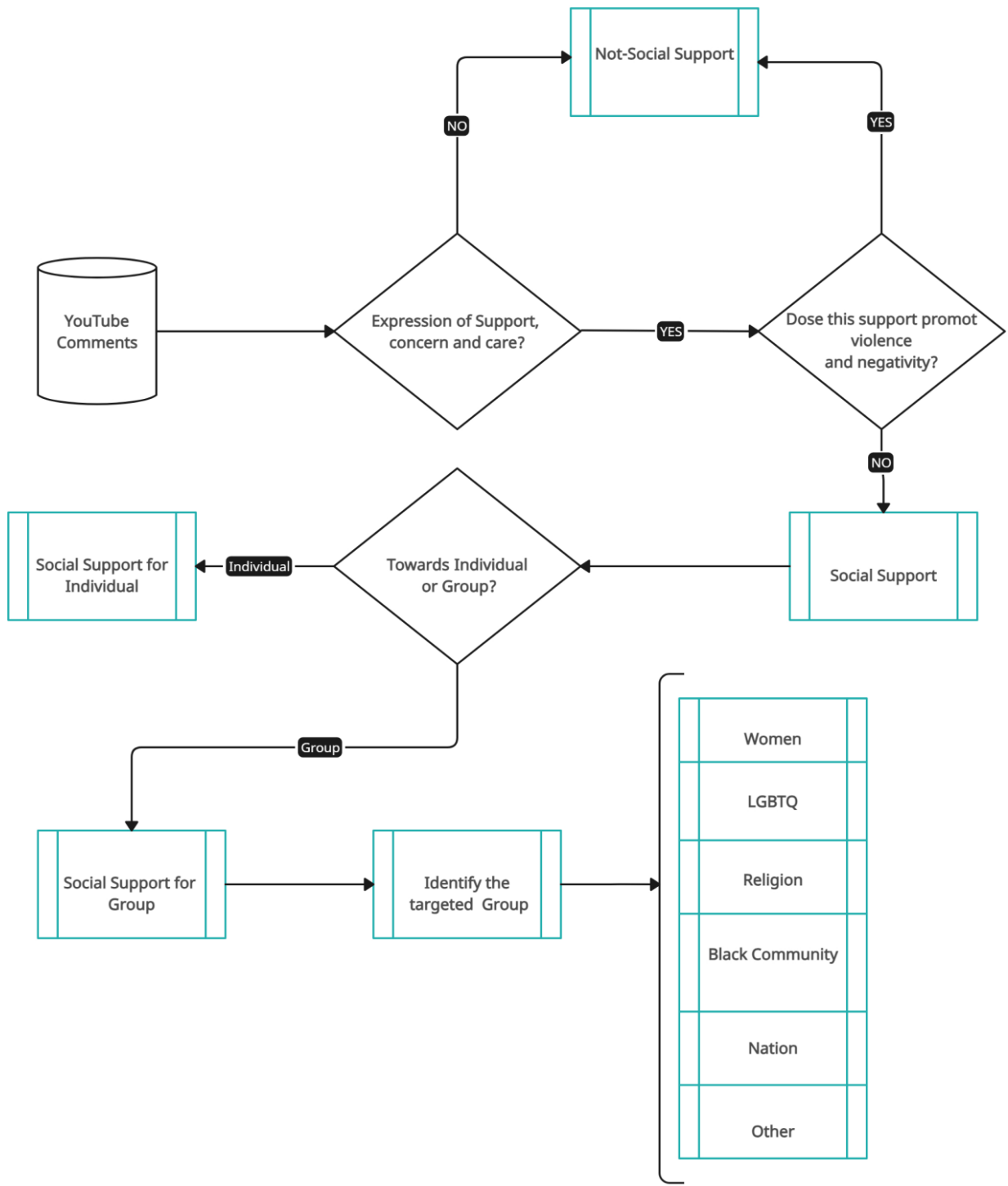

Figure 1: Overview of annotation procedure

*5.1. LIWC*

The LIWC model has transformed psychological research by making language data analysis more robust, accessible, and scientifically rigorous. LIWC-22 evaluates over 100 textual dimensions, all validated by esteemed research institutions worldwide. With over 20,000 scientific publications utilizing LIWC, it has become a widely recognized and trusted tool, enabling novel analytical approaches. Despite its advantages, LIWC has limitations. One issue is its reliance on predefined linguistic categories, which may not capture the nuances and variations of natural language (Lyu et al., 2023; Bojić, 2023). Additionally, LIWC can struggle with accurately interpreting sarcasm, irony, and other subtle language forms, potentially leading to misinterpretations. In recent years, machine learning and LIWC have risen as formidable assets in the realm of Psychology. Their advent has brought about groundbreaking advancements in mental disorder diagnosis, while also making substantial contributions to enhancing human well-being (Bahar and Ülker). LIWC stands as a widely embraced computerized text analysis software, facilitating researchers in scrutinizing the emotional, cognitive, structural, and process components inherent in language.

This is achieved through the frequency analysis of words in written text or speech (Thompson and Hartwig, 2023). Recent studies suggest an increasing trend in utilizing both LIWC and machine learning for diagnosing mental disorders. Research has concentrated on analyzing large volumes of text to uncover links between everyday language usage and traits such as personality, social interactions, and cognitive patterns (Bahar and Ülker).

In this study, we aim to employ machine learning and LIWC to detect social support. We employed a wide-ranging set of exit analysis features, encompassing Summary Variables, Linguistic Dimensions, Drives, Cognition, Affect, Social Processes, Culture, Lifestyle, Physical Attributes, States, Motives, Perception, and Conversation extracted from LIWC. The average statistical relationships between these features and various types of social supportive are detailed in Table 4.

**Word count** Word Count (WC) measures user engagement and fluency by evaluating the diversity and range of vocabulary used. It reveals that supportive interactions generally have higher word counts than non-supportive ones, suggesting that people give more detailed responses when being supportive. This trend is consistent in both individual and group settings. Additionally, within group interactions, there is a tendency to provide more detailed responses when offering support to women.

The **Function words** analyzed in LIWC include pronouns, impersonal pronouns, articles, prepositions, auxiliary verbs, common adverbs, conjunctions, and negations. Pronouns, in particular, provide insights into users' personalities and communication styles. Supportive comments use fewer function words than non-supportive ones, emphasizing sentiment over structure. Individual comments also use fewer function words than group comments, indicating more direct communication. Discussions on LGBTQ topics show higher usage of function words, reflecting the complexity and sensitivity of the subject and the diverse perspectives within the community.

**Drives** encompass broad dimensions of needs and motivations. In the "Drive" category, supportive comments are more frequent than non-supportive ones, indicating a tendency towards encouragement in such discussions. The difference between group and individual comments is minimal, showing balanced contributions. However, within the broader "Group" category, the "Other" subgroup has a notably higher value, suggesting a significant presence of miscellaneous topics. This diversity reflects unique viewpoints or niche topics, adding complexity and richness to group discussions.

**Cognition** sheds light on the writer's reasoning and thought processes. LIWC analysis shows that supportive comments contain more cognitive content than non-supportive ones, highlighting thoughtful and empathetic responses. Group discussions exhibit higher cognitive engagement than individual comments, indicating that group dynamics encourage deeper analysis. Additionally, the LGBTQ subgroup within group discussions shows elevated cognitive focus, suggesting heightened awareness and contemplation of LGBTQ issues, leading to more informed and nuanced discourse.

**Affect** includes various emotional aspects like positive and negative emotions, anxiety, anger, sadness, and swear words. In the "Affect" category, supportive comments show more emotional expression than non-supportive ones, reflecting empathy and encouragement. Individual comments exhibit higher emotional expression than group discussions, indicating more intense personal reflections. However, the "black community" subgroup shows lower emotional expression, suggesting a need for greater sensitivity and empathy in discussions about Black individuals. This disparity highlights potential differences in emotional engagement across demographic and thematic contexts in online discourse.

The LIWC analysis of **Social Processes** highlights key dynamics in online interactions. Supportive comments emphasize social processes more than non-supportive ones, indicating efforts to build connections and offer assistance. Individuals show higher social engagement than groups, suggesting a focus on interpersonal connections in one-on-one interactions. The "religion" subgroup has a lower value in group interactions, pointing to a relative scarcity of discussions on religious topics. This might indicate varying comfort levels or relevance in discussing religion online, underscoring the need for sensitivity and inclusivity regarding religious diversity in online discourse.

The **Culture** category covers Politics, Ethnicity, and Technology, focusing on political discourse, ethnic identities, and scientific advancements. LIWC analysis shows that supportive comments are more prevalent in cultural discussions, indicating positivity and understanding. Group discussions demonstrate higher cultural engagement than individual comments, suggesting collective exploration of cultural topics. The significant elevation of the "religion" subgroup highlights the importance of religious themes within cultural discourse, emphasizing the need to understand and respect diverse religious perspectives in cultural dialogue.

| Feature | Social Support | Not Social Support | Group | Individual | Nation | Religion | Black Community | Women | Other | LGBTQ |
|---|---|---|---|---|---|---|---|---|---|---|
| **Word count** | 44.0645 | 41.5249 | 44.0346 | 44.1927 | 43.9855 | 46.7228 | 45.1554 | 49.7326 | 42.6865 | 46.8500 |
| **Function words** | 11.7344 | 12.4165 | 11.5321 | 12.6014 | 10.9675 | 10.8692 | 11.8174 | 12.5538 | 11.8160 | 13.8851 |
| **Drives** | 3.0849 | 2.8181 | 3.0875 | 3.0736 | 3.1422 | 2.9367 | 2.8167 | 3.1643 | 3.3258 | 2.1412 |
| **Cognition** | 3.2936 | 4.1433 | 3.3058 | 3.2413 | 2.6595 | 3.2193 | 4.7268 | 3.6753 | 3.6126 | 5.2924 |
| **Affect** | 3.3550 | 2.1917 | 3.207 | 3.9894 | 3.3130 | 2.6355 | 1.7190 | 2.0784 | 3.4503 | 3.0571 |
| **Social processes** | 4.1968 | 3.6111 | 3.8995 | 5.4712 | 3.6828 | 3.3379 | 3.3794 | 4.7686 | 4.5102 | 3.5380 |
| **Culture** | 1.1013 | 0.9521 | 1.2687 | 0.3841 | 1.5139 | 4.4231 | 2.7719 | 0.3347 | 0.7258 | 0.1812 |
| **Lifestyle** | 1.6341 | 1.1859 | 1.5007 | 2.2061 | 1.6428 | 5.0315 | 0.5 | 0.6790 | 1.6762 | 0.4349 |
| **Physical** | 0.4253 | 0.3663 | 0.4414 | 0.3564 | 0.3394 | 0.1645 | 0.5611 | 0.2457 | 0.4902 | 0.9029 |
| **States** | 0.2726 | 0.3464 | 0.2803 | 0.2394 | 0.2436 | 0.1080 | 0.3325 | 0.5148 | 0.3219 | 0.3204 |
| **Motives** | 2.5244 | 2.1321 | 2.4793 | 2.7175 | 2.6755 | 1.8255 | 1.5546 | 2.2067 | 2.3508 | 2.4705 |
| **Perception** | 2.5254 | 2.8775 | 2.5355 | 2.4818 | 2.3648 | 1.9381 | 3.6379 | 2.8534 | 2.5619 | 2.7431 |
| **Conversation** | 0.4950 | 0.5146 | 0.4795 | 0.5616 | 0.5037 | 0.2757 | 0.4729 | 0.308 | 0.4431 | 0.5049 |

Table 4: lIWC

The **Lifestyle** category includes work, home life, school, and employment. LIWC analysis reveals that supportive comments are more common in lifestyle discussions, indicating positivity and encouragement. Individual comments show higher engagement than group discussions, suggesting that personal perspectives and experiences are more frequently shared in lifestyle topics. The significant elevation of the "religion" subgroup within this category highlights the strong influence of religious beliefs and values on individuals' lifestyles, emphasizing the need to respect diverse religious perspectives in lifestyle discourse.

The **Physical** dimension includes health-related terms in subcategories like illness, wellness, mental health, and substances. LIWC analysis shows that supportive comments are more prevalent in physical discussions, indicating positivity and empathy. Group discussions show higher engagement with physical topics compared to individual comments, suggesting collective exploration. The lower value of the "religion" subgroup in this category implies fewer discussions on physical aspects within religious discourse, indicating that physical topics may not be as central in religious discussions, highlighting differing focuses across thematic contexts in online discourse.

**States** encompass transient internal conditions that can influence behaviors. In the "State" category, non-supportive comments and group interactions are more prevalent, indicating a focus on expressing or responding to various emotional states within group dynamics. The elevated value of the "Women" subgroup suggests a heightened discussion of women's internal states and experiences, reflecting an awareness of gender-specific perspectives and issues in online discourse about internal conditions and behaviors.

**Motives** are fundamental internal states that drive, direct or attract an individual's behavior (Boyd et al., 2022). In the "Motives" category, the LIWC analy. LIWC analysis reveals that in the "Motives" category, supportive comments, individual perspectives, and discussions about national identity are more prevalent. This suggests that online discourse often features expressions of empathy, encouragement, personal aspirations, and patriotism, reflecting a strong connection to collective identity and societal values. These findings highlight the complex interplay between individual, interpersonal, and societal motivations in digital interactions, showcasing the diverse factors that drive human behavior and expression online.

**Perception** involves elements such as attention, motion, spatial awareness, visual and auditory stimuli, and emotional responses Boyd et al. (2022). LIWC analysis shows that in the "Perception" category, non-supportive comments, group interactions, and discussions about Black individuals or issues are more common. This suggests a focus on judgment, interpretation, or perspective in non-supportive and group dynamics. The prominence of perception-related content in discussions about Black individuals underscores the importance of understanding perception and misperception in conversations about race and racial identity. These findings highlight the critical role of perception in interpersonal interactions and social issue discussions.

**Conversation** includes elements such as Netspeak, Assent, Nonfluencies, and Fillers (Boyd et al., 2022). LIWC analysis shows that in the "Conversation" category, non-supportive comments, individual perspectives, and LGBTQ-related discussions are particularly prevalent. This indicates that online interactions often feature conflict or disagreement, personal expression, and significant focus on LGBTQ topics. The prevalence of non-supportive language highlights the presence of disagreement in digital conversations, while individual perspectives underscore personal expression and autonomy. Additionally, the emphasis on LGBTQ-related discourse reflects the importance of discussions on sexual orientation, gender identity, and LGBTQ rights. These findings highlight the diverse conversational dynamics in digital discourse, covering both interpersonal interactions and social identity topics.

*5.2. Emotions*

This study employed the NRC Emotion Lexicon (Mohammad and Turney, 2013) to analyze emotions associated with different types of social support. Table 5 presents emotions categorized into supportive and non-supportive groups. It shown that certain emotions such as joy and trust generally have higher intensity levels in supportive contexts compared to non-supportive ones. For instance, joy tends to be significantly higher in supportive contexts across various categories like individual, group, nation, and others. Conversely, emotions like anger and fear show more nuanced patterns, with their intensity varying across different categories. Additionally, there are notable disparities in emotion intensity across different demographic groups, such as LGBTQ individuals, black community, women, and religious groups. These variations may reflect the complex interplay between emotions and social dynamics within different contexts and communities.

*5.3. Sentiment Analysis*

The Social Support dataset was analyzed for sentiment using VaderSentiment (Hutto and Gilbert, 2014). Table 6 showcases the distribution of sentiment (negative, neutral, and positive) within different categories. Across most categories, neutral sentiment appears to be the most prevalent, followed by either positive or negative sentiment, though the balance between the two varies. Non-supportive contexts generally exhibit slightly higher proportions of negative sentiment compared to supportive contexts, where neutral sentiment tends to dominate. Interestingly, individual and group contexts show a relatively balanced distribution between neutral and positive sentiments. There are also noticeable differences in sentiment distribution across demographic groups, with variations particularly evident among black community and women, where negative sentiment appears more pronounced. This suggests that sentiment dynamics may be influenced by factors such as social context and demographic characteristics.

## 6. Experiments

*6.1. Traditional machine learning models*

Five traditional machine learning classifiers, including LR and SVM with both radial basis function (RBF) and linear kernels, DT, and RFC, are employed for the task of detecting hope speech. We experimented with their soft and hard voting as ensemble methods to create more robust models. All classifiers are utilized with their default parameters and trained on the TF-IDF of unigrams. Additionally, features derived from LIWC, emotion analysis, and sentiment analysis are incorporated to study their contribution to the classification task in the proposed dataset.

*6.2. Preprocessing*

Initially, the data preprocessing involved the removal of duplicate comments and the selection of English tweets. Following this, tokenization, lowercasing, punctuation removal, stop word elimination, and stemming or lemmatization were performed to standardize the text data. Additionally, emojis and emoticons were converted into textual representations using the emotion library Mohammad and Turney (2013). Subsequently, abbreviations were expanded to their full forms utilizing a predefined dictionary, while punctuation marks and stopwords were further removed to refine the text data.

*6.3. Feature extraction*

Psycholinguistic, emotional, and sentiment attributes were extracted from the textual corpus through the utilization of established lexicons and sentiment analysis tools, namely LIWC by Tausczik and Pennebaker (2010), the NRC Emotion Lexicon devised by Mohammad and Turney (2013), and VaderSentiment introduced by Hutto and Gilbert (2014)Hutto and Gilbert (2014). These metrics were employed to construct comprehensive feature vectors for each text, encapsulating the psychological, emotional, and sentiment dimensions inherent in every tweet. Furthermore, to enhance the scope of analysis and facilitate comparative assessments across different feature sets, n-gram features were also derived. The quantification of the feature space dimensions is elaborated upon in detail within Table 9. module.

*6.4. Model training and predictions*

In all experiments, we utilized a 5-fold cross-validation approach for both training and evaluating the ML models. Evaluation and comparison were conducted based on the average weighted and macro scores across all folds. Comprehensive results are elaborated upon in the Results section (7).

*6.5. Deep learning*

Two deep learning models, namely Convolutional Neural Network (CNN) and Bidirectional Long Short-Term Memory (BiLSTM), were trained separately using Global Vectors for Word Representation (GloVe) and FastText embeddings. A Keras tokenizer was fitted on the dataset texts to convert all texts into sequences. The maximum sequence length was set to the maximum length of tweets in the dataset, and all sequences were padded to this length. Vectors were obtained from the word embedding matrix for each tweet, after which the input sequences were created and fed to the deep learning models. The parameters used for both models are detailed in Table 7, and the models were trained for 50 epochs for each fold.

## 7. Results

The machine learning models were rigorously evaluated across three distinct classification steps as discussed in section 4.3. Importantly, each level of our experiment was augmented with different feature combinations, namely LIWC+Emotions and sentiment features only, TF-IDF only, and a combination of all features. This meticulous approach allowed us to systematically investigate the impact of different feature sets on the performance of various models across different classification tasks. Through this comprehensive analysis, we aimed to identify the most effective model-feature combinations for accurate and reliable social support detection. We conducted experiments using CNN and BiLSTM models with GloVe and FastText embeddings across three subtasks, with results detailed in Section 7.3.1.

| Subtasks | Labels | Anger | Anticipation | Disgust | Fear | Joy | Sadness | Surprise | Trust |
|---|---|---|---|---|---|---|---|---|---|
| Subtask1 | Social Support | 0.6399 | 1.2464 | 0.5223 | 1.0840 | 1.6923 | 1.0035 | 0.4190 | 1.6265 |
| | Not Social Support | 0.6946 | 0.9738 | 0.4987 | 0.9263 | 1.0332 | 0.8737 | 0.4408 | 1.2711 |
| Subtask2 | Individual | 0.4066 | 1.6548 | 0.4609 | 0.8723 | 2.1583 | 0.6595 | 0.4893 | 1.8983 |
| | Group | 0.6944 | 1.1511 | 0.5366 | 1.1334 | 1.5835 | 1.0838 | 0.4026 | 1.5631 |
| Subtask3 | Nation | 0.5651 | 0.8788 | 0.4327 | 0.9501 | 1.2932 | 0.8156 | 0.3319 | 1.2219 |
| | Other | 0.6596 | 1.3807 | 0.5057 | 1.2442 | 1.7615 | 1.0730 | 0.4211 | 1.8307 |
| | LGBTQ | 0.9675 | 1.5519 | 0.9935 | 1.2922 | 2.7402 | 1.1818 | 0.6623 | 1.8701 |
| | Black Community | 1.4824 | 1.4736 | 0.8508 | 1.8508 | 1.4736 | 3.2894 | 0.4912 | 2.3947 |
| | Religion | 0.8947 | 2.6315 | 0.6315 | 1.6315 | 2.6842 | 0.9473 | 0.5263 | 3.6315 |
| | Women | 1.0833 | 2.0416 | 0.9583 | 1.4166 | 1.8333 | 1.2916 | 0.7083 | 2.1666 |

Table 5: Emotions

| Subtasks | Labels | Negative | Neutral | Positive |
|---|---|---|---|---|
| Subtask1 | Social Support | 0.1498 | 0.4702 | 0.3798 |
| | Not Social Support | 0.1747 | 0.5741 | 0.2511 |
| Subtask2 | Individual | 0.1183 | 0.4233 | 0.4583 |
| | Group | 0.1572 | 0.4812 | 0.3615 |
| Subtask3 | Nation | 0.1360 | 0.4830 | 0.3809 |
| | Other | 0.1990 | 0.4497 | 0.3511 |
| | LGBTQ | 0.1171 | 0.4880 | 0.3947 |
| | Black Community | 0.2070 | 0.5780 | 0.2149 |
| | Religion | 0.1176 | 0.4826 | 0.3996 |
| | Women | 0.1714 | 0.5844 | 0.2441 |

Table 6: Sentiment Analysis

Table 7: Parameters for deep learning models.

| Parameters | CNN | BiLSTM |
|---|---|---|
| Epochs | 50 per fold | 50 per fold |
| Optimizer | Adam | Adam |
| Loss | categorical crossentropy | categorical crossentropy |
| Embedding size | 300 | 300 |
| Learning rate (lr) | 0.001 | 0.001 |
| Dropout | 0.1 | 0.1 |
| Activation | softmax | softmax |

*7.1. Social Support Detection with LIWC, emotions, and sentiments features*

In Table 8, where we incorporated LIWC, emotions, and sentiment analysis as features, our analysis delved into different traditional machine learning models across various levels of our experiment. Notably, in subtask 1, the LR model emerges as the frontrunner, boasting a macro F1 score of 0.7061. This trend persists in subtask 2, where LR again outshines other models with an impressive macro F1 score of 0.7751. Extending this trend, LR also leads the pack in the subtask 3, securing a macro F1 score of 0.5666. These consistent superior performances by the LR model underscore its efficacy across different levels of our experiment. The LR model's strengths in maintaining competitive macro F1 scores across diverse classification tasks suggest its adaptability and robustness in capturing nuanced patterns within the data. This reaffirms the importance of leveraging LR's simplicity and interpretability to achieve reliable and consistent results in social support detection tasks.

*7.2. Social Support Detection using Unigram with TF-IDF values*

In this section, we conducted our experiment using Unigram with TF-IDF values, yielding distinct outcomes compared to the previous section. Table 9 showcases notable variations in model performances across different classification tasks. For subtask 1, the Soft Voting model emerges as the frontrunner, surpassing other models in terms of performance metrics. Similarly, in subtask 2, the Soft Voting model exhibits superior results compared to its counterparts. This trend persists in subtask 3, where, once again, the Soft Voting model demonstrates the highest performance. These findings underscore the efficacy of Soft Voting in leveraging Unigram with TF-IDF values for social support detection across various classification tasks. The superior performance of the Soft Voting ensemble model across all classification tasks could be attributed to its ability to aggregate predictions from multiple base models in a balanced manner. Soft Voting combines the predictions of individual models by taking into account their probabilities rather than simple majority voting, thus leveraging the collective wisdom of diverse classifiers. This ensemble approach often leads to more robust and generalized predictions, as it mitigates the biases and weaknesses inherent in individual models. Additionally, Soft Voting can effectively exploit the complementary strengths of different base models, resulting in enhanced overall performance

*7.3. Social Support Detecion with the combination of all features*

In this section, we integrated a combination of all features including LIWC, emotions, sentiment scores, and Unigram with TF-IDF values. Table 10 presents the outcomes of our experiments across different classification tasks. In the first subtask, SVM (Linear) emerges as the top-performing model, surpassing others in terms of performance metrics. Moving to the second subtask, soft voting stands out with the highest performance value among the models considered. Transitioning to the third subtask, hard voting demonstrates superior results compared to other models. These findings highlight the varied strengths of different models when leveraging a comprehensive feature set, underscoring the importance of selecting appropriate models tailored to the specific task requirements.

| Models | Avg. weighted scores | | | Avg. macro scores | | | Accuracy |
|---|---|---|---|---|---|---|---|
| | Precision | Recall | F1-score | Precision | Recall | F1-score | |
| | Subtask1 | | | | | | |
| **LR** | **0.8151** | **0.8286** | **0.8111** | **0.7737** | **0.6794** | **0.7061** | **0.8286** |
| SVM(rbf) | 0.8137 | 0.8154 | 0.7733 | 0.8095 | 0.6083 | 0.6265 | 0.8154 |
| SVM (linear) | 0.8179 | 0.8297 | 0.8080 | 0.7873 | 0.6688 | 0.6977 | 0.8297 |
| DT | 0.7512 | 0.7487 | 0.7497 | 0.6402 | 0.6428 | 0.6412 | 0.7487 |
| RFC | 0.8229 | 0.8288 | 0.8000 | 0.8098 | 0.6492 | 0.6785 | 0.8288 |
| Soft voting | 0.8229 | 0.8388 | 0.8066 | 0.7866 | 0.6721 | 0.7033 | 0.8312 |
| Hard voting | 0.8267 | 0.8326 | 0.8064 | 0.8124 | 0.6597 | 0.6905 | 0.8326 |
| | Subtask2 | | | | | | |
| **LR** | **0.8680** | **0.8756** | **0.8688** | **0.8130** | **0.7506** | **0.7751** | **0.8756** |
| SVM(rbf) | 0.8443 | 0.8403 | 0.7952 | 0.8535 | 0.5887 | 0.6066 | 0.8403 |
| SVM (linear) | 0.8658 | 0.8729 | 0.8661 | 0.8095 | 0.7469 | 0.7709 | 0.8729 |
| DT | 0.8030 | 0.7996 | 0.8010 | 0.6759 | 0.6800 | 0.6774 | 0.7996 |
| RFC | 0.8646 | 0.8698 | 0.8521 | 0.8435 | 0.6940 | 0.7335 | 0.8698 |
| Soft voting | 0.8648 | 0.8729 | 0.8629 | 0.8191 | 0.7306 | 0.7613 | 0.8729 |
| Hard voting | 0.8688 | 0.8752 | 0.8620 | 0.8380 | 0.7195 | 0.7558 | 0.8752 |
| | Subtask3 | | | | | | |
| **LR** | **0.7056** | **0.7010** | **0.7000** | **0.5666** | **0.5940** | **0.5666** | **0.7010** |
| SVM(rbf) | 0.5807 | 0.6321 | 0.5801 | 0.3355 | 0.2974 | 0.2973 | 0.6321 |
| SVM (linear) | 0.6844 | 0.6784 | 0.6773 | 0.5269 | 0.5606 | 0.5303 | 0.6784 |
| DT | 0.6450 | 0.6348 | 0.6358 | 0.4757 | 0.4851 | 0.4703 | 0.6348 |
| RFC | 0.7149 | 0.7308 | 0.7031 | 0.5222 | 0.4302 | 0.4500 | 0.7308 |
| Soft Voting | 0.7297 | 0.7369 | 0.7241 | 0.6689 | 0.5082 | 0.5490 | 0.7369 |
| Hard voting | 0.7167 | 0.7275 | 0.7082 | 0.6221 | 0.5011 | 0.5234 | 0.7275 |

Table 8: Results using LIWC, emotions and sentiments features

| Models | Avg. weighted scores | | | Avg. macro scores | | | Accuracy |
|---|---|---|---|---|---|---|---|
| | Precision | Recall | F1-score | Precision | Recall | F1-score | |
| | Subtask1 | | | | | | |
| LR | 0.8547 | 0.8608 | 0.8491 | 0.8329 | 0.7364 | 0.7682 | 0.8608 |
| SVM(rbf) | 0.8530 | 0.8591 | 0.8466 | 0.8324 | 0.7311 | 0.7636 | 0.8591 |
| SVM (linear) | 0.8516 | 0.8589 | 0.8493 | 0.8207 | 0.7432 | 0.7707 | 0.8589 |
| DT | 0.8063 | 0.8092 | 0.8076 | 0.7246 | 0.7172 | 0.7206 | 0.8092 |
| RFC | 0.8463 | 0.8536 | 0.8405 | 0.8213 | 0.7227 | 0.7539 | 0.8536 |
| **Soft voting** | **0.8543** | **0.8611** | **0.8512** | **0.8265** | **0.7447** | **0.7733** | **0.8611** |
| Hard voting | 0.8541 | 0.8602 | 0.8482 | 0.8326 | 0.7348 | 0.7667 | 0.8602 |
| | Subtask2 | | | | | | |
| LR | 0.8604 | 0.8622 | 0.8362 | 0.8554 | 0.6589 | 0.6977 | 0.8622 |
| SVM(rbf) | 0.8647 | 0.8698 | 0.8503 | 0.8490 | 0.6877 | 0.7286 | 0.8698 |
| SVM (linear) | 0.8632 | 0.8716 | 0.8615 | 0.8184 | 0.7281 | 0.7594 | 0.8716 |
| DT | 0.8346 | 0.8376 | 0.8358 | 0.7354 | 0.7243 | 0.7292 | 0.8376 |
| RFC | 0.8653 | 0.8716 | 0.8559 | 0.8412 | 0.7037 | 0.7426 | 0.8716 |
| **Soft voting** | **0.8704** | **0.8779** | **0.8689** | **0.8279** | **0.7423** | **0.7730** | **0.8779** |
| Hard voting | 0.8641 | 0.8707 | 0.8537 | 0.8410 | 0.6983 | 0.7376 | 0.8707 |
| | Subtask3 | | | | | | |
| LR | 0.7848 | 0.7909 | 0.7816 | 0.6453 | 0.5192 | 0.5486 | 0.7909 |
| SVM(rbf) | 0.7837 | 0.7915 | 0.7824 | 0.6467 | 0.5454 | 0.5708 | 0.7915 |
| SVM (linear) | 0.7995 | 0.8064 | 0.8010 | 0.6936 | 0.6275 | 0.6458 | 0.8064 |
| DT | 0.7929 | 0.7876 | 0.7873 | 0.7105 | 0.6864 | 0.6807 | 0.7876 |
| RFC | 0.8249 | 0.8290 | 0.8227 | 0.7245 | 0.6522 | 0.6667 | 0.8290 |
| **Soft voting** | **0.8280** | **0.8273** | **0.8239** | **0.8074** | **0.7043** | **0.7262** | **0.8273** |
| Hard voting | 0.8051 | 0.8102 | 0.8033 | 0.7202 | 0.6046 | 0.6306 | 0.8102 |

Table 9: Results using Unigram with TF-IDF values

| Models | Avg. weighted scores | | | Avg. macro scores | | | Accuracy |
|---|---|---|---|---|---|---|---|
| | Precision | Recall | F1-score | Precision | Recall | F1-score | |
| | Subtask1 | | | | | | |
| LR | 0.8175 | 0.8306 | 0.8145 | 0.7753 | 0.6862 | 0.7127 | 0.8306 |
| SVM(rbf) | 0.8115 | 0.8134 | 0.7696 | 0.8072 | 0.6031 | 0.6194 | 0.8134 |
| SVM (linear) | **0.8558** | **0.8626** | **0.8557** | **0.8190** | **0.7602** | **0.7830** | **0.8626** |
| DT | 0.7889 | 0.7905 | 0.7897 | 0.6974 | 0.6946 | 0.6959 | 0.7905 |
| RFC | 0.8465 | 0.8492 | 0.8292 | 0.8391 | 0.6944 | 0.7303 | 0.8492 |
| Soft voting | 0.8500 | 0.8565 | 0.8433 | 0.8285 | 0.7256 | 0.7579 | 0.8565 |
| Hard voting | 0.8534 | 0.8576 | 0.8424 | 0.8407 | 0.7189 | 0.7544 | 0.8576 |
| | Subtask2 | | | | | | |
| LR | 0.8678 | 0.8752 | 0.8684 | 0.8126 | 0.7505 | 0.7747 | 0.8752 |
| SVM(rbf) | 0.8497 | 0.8398 | 0.7924 | 0.8699 | 0.5841 | 0.5997 | 0.8398 |
| SVM (linear) | **0.8783** | **0.8828** | **0.8792** | **0.8186** | **0.7808** | **0.7969** | **0.8828** |
| DT | 0.8416 | 0.8465 | 0.8437 | 0.7498 | 0.7316 | 0.7399 | 0.8465 |
| RFC | 0.8709 | 0.8698 | 0.8472 | 0.8731 | 0.6774 | 0.7204 | 0.8698 |
| Soft voting | 0.8828 | 0.8886 | 0.8798 | 0.8528 | 0.7561 | 0.7910 | 0.8886 |
| Hard voting | 0.8791 | 0.8832 | 0.8702 | 0.8614 | 0.7285 | 0.7696 | 0.8832 |
| | Subtask3 | | | | | | |
| LR | 0.7116 | 0.7120 | 0.7078 | 0.6126 | 0.5720 | 0.5717 | 0.7120 |
| SVM(rbf) | 0.5837 | 0.6310 | 0.5740 | 0.3476 | 0.2948 | 0.2927 | 0.6310 |
| SVM (linear) | 0.7260 | 0.7186 | 0.7193 | 0.5466 | 0.5746 | 0.5482 | 0.7186 |
| DT | 0.7698 | 0.7677 | 0.7670 | 0.6465 | 0.6448 | 0.6348 | 0.7677 |
| RFC | 0.7620 | 0.7793 | 0.7590 | 0.5392 | 0.5024 | 0.5145 | 0.7793 |
| Soft Voting | 0.7736 | 0.7848 | 0.7703 | 0.6375 | 0.5711 | 0.5877 | 0.7848 |
| Hard voting | **0.7980** | **0.8041** | **0.7959** | **0.7248** | **0.5998** | **0.6355** | **0.8041** |

Table 10: Results using all features

In Table 10, the comparison across three levels of SSD reveals the performance of various models with different feature combinations. In the first subtask, the combination of LIWC, Emotions and sentiments features, and Unigram with TF-IDF values with the SVM (linear) model demonstrate superior performance. Moving to the second subtask, we observe that the combination of LIWC, Emotions and sentiment analysis, and Unigram with TF-IDF values with the soft voting model yields the highest value. Finally, in the third subtask, the TF-IDF feature with the soft voting model exhibits the highest performance. These findings underscore the influence of both feature set and model choice on the effectiveness of SSD across different subtasks.

*7.3.1. Deep learning*

Table 11 provides an overview of model performance across various tasks, with a focus on the macro F1-score, which is considered a robust metric for evaluating model performance in tasks with imbalanced classes. Across different word embeddings (GloVe and FastText) and model architectures (CNN and BiLSTM), the configurations yielding the highest macro F1-scores are of particular interest. For example, in Task1, using FastText embeddings with the BiLSTM model resulted in a macro F1-score of 0.7611, indicating a balanced performance in terms of precision and recall across both support and non-support classes. Similarly, in Task 2 and Task 3, certain configurations achieved macro F1-scores of 0.8184 and 0.7235 respectively, suggesting strong overall performance across all classes.

**8. Error analysis**

The detailed results of all experiments are presented in Section 7. Here, we analyze the performance of the best-performing model for each subtask.

Table 12 presents the classwise scores for the best-performing models. For Subtask 1 and Subtask 2, the CNN with a Glove embedding. In Subtask 3, the best performance was achieved through soft-voting of models using only Unigram with TF-IDF values.

Comparing the results for each category with the label distribution across the dataset in Table 3 reveals the influence of these imbalanced distributions on performance. In all subtasks, categories with a higher population were predicted more accurately.

We also provided the confusion matrices for these models in Figures 2, 3, and 4. The analysis of these confusion matrices reveals several key patterns of misclassification.

For subtask 1, supportive comments were frequently misclassified as not supportive. This indicates that the model struggles to accurately distinguish between these two categories, likely due to subtle differences in language or context that are challenging for the algorithm to capture.

Table 11: Results for deep learning models

| Word Embedding | Tasks | Model | Weighted scores | | | Macro scores | | | Accuracy |
|---|---|---|---|---|---|---|---|---|---|
| | | | precision | recall | F1-score | precision | recall | F1-score | |
| **GloVe** | Task1 | CNN | 0.8406 | 0.8449 | 0.8355 | 0.7819 | 0.7520 | 0.7526 | 0.8449 |
| | | BiLSTM | **0.8392** | **0.8448** | **0.8389** | **0.7751** | **0.7587** | **0.7611** | **0.8448** |
| **FastText** | | CNN | 0.8007 | 0.8099 | 0.8009 | 0.7231 | 0.6984 | 0.7028 | 0.8099 |
| | | BiLSTM | 0.8130 | 0.8122 | 0.8082 | 0.7280 | 0.7287 | 0.7204 | 0.8122 |
| **GloVe** | Task2 | CNN | **0.8913** | **0.8957** | **0.8921** | **0.8444** | **0.7988** | **0.8184** | **0.8957** |
| | | BiLSTM | 0.8816 | 0.8836 | 0.8825 | 0.8118 | 0.8002 | 0.8058 | 0.8836 |
| **FastText** | | CNN | 0.8444 | 0.8524 | 0.8471 | 0.7633 | 0.7277 | 0.7426 | 0.8524 |
| | | BiLSTM | 0.8521 | 0.8569 | 0.8538 | 0.7704 | 0.7467 | 0.7571 | 0.8569 |
| **GloVe** | Task3 | CNN | 0.8593 | 0.8576 | 0.8538 | 0.7890 | 0.6981 | 0.7143 | 0.8576 |
| | | BiLSTM | **0.8445** | **0.8378** | **0.8373** | **0.7595** | **0.7201** | **0.7235** | **0.8378** |
| **FastText** | | CNN | 0.7758 | 0.7788 | 0.7717 | 0.6791 | 0.5813 | 0.6055 | 0.7788 |
| | | BiLSTM | 0.7587 | 0.7600 | 0.7550 | 0.5937 | 0.5686 | 0.5664 | 0.7600 |

In subtask 2, there was a notable confusion between support for individuals and support for groups. This suggests that the model finds it difficult to differentiate when the support is directed towards a single person versus a collective group, possibly because the expressions of support can be quite similar in both scenarios.

In subtask 3, when predicting targeted groups, the model often misclassified religious groups as nations. This specific confusion was more prominent than any other misclassification within this subtask. It indicates a particular challenge in distinguishing between religion-based and nation-based supportive comments, which may stem from overlapping cultural or contextual cues in the data.

Additionally, across all subtasks, there was significant misclassification into the "other" category. This category encompasses instances that may target multiple groups or do not fit neatly into predefined categories, making it a frequent source of confusion for the model.

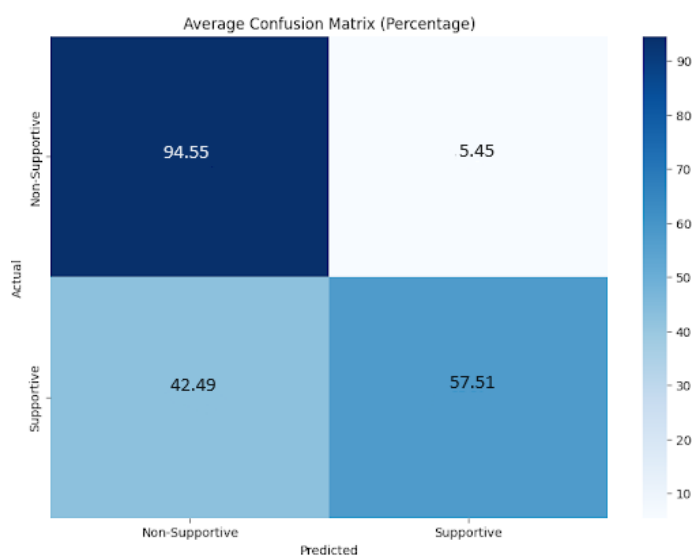

Figure 2: Confusion matrix for the best-performing model in subtask 1 (SVM (linear) + all features)

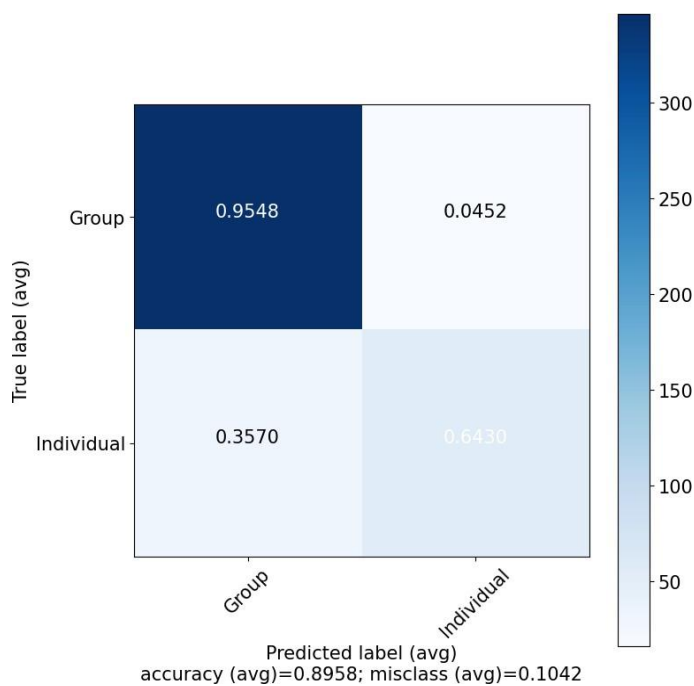

Figure 3: Confusion matrix for the best-performing model in subtask 2 (CNN + Glove)

## 9. Discussion

The dataset and experiments proposed in this paper have several characteristics and limitations that are briefly discussed in the following:

- The dataset utilized in this paper was exclusively gathered from YouTube comments within a specific timeframe and targeted videos, which may introduce biases. To enhance the diversity dataset and its richness, incorporating posts from other social media platforms, such as X and Reddit, in an open timeframe would be beneficial.

| Tasks | Model | Feature set | Label | precision | recall | F1-score |
|---|---|---|---|---|---|---|
| Subtask1 | SVM (Linear) | All Feature | Social Support | 0.7528 | 0.5750 | 0.6517 |
| | | | Non-Social Support | 0.8854 | 0.9455 | 0.9144 |
| Subtask2 | CNN | Glove | Group | 0.9198 | 0.9547 | 0.9369 |
| | | | Individual | 0.7691 | 0.6429 | 0.6999 |
| Subtask3 | Soft voting | TF_IDF | Black Community | 0.8438 | 0.8684 | 0.8546 |
| | | | LGBTQ | 0.9199 | 0.8856 | 0.9012 |
| | | | Other | 0.7270 | 0.7142 | 0.7193 |
| | | | Religion | 0.7333 | 0.3452 | 0.4233 |
| | | | Women | 0.7600 | 0.5214 | 0.5835 |
| | | | Nation | 0.8610 | 0.8913 | 0.8755 |

Table 12: Classwise scores for best performing model for each subtask

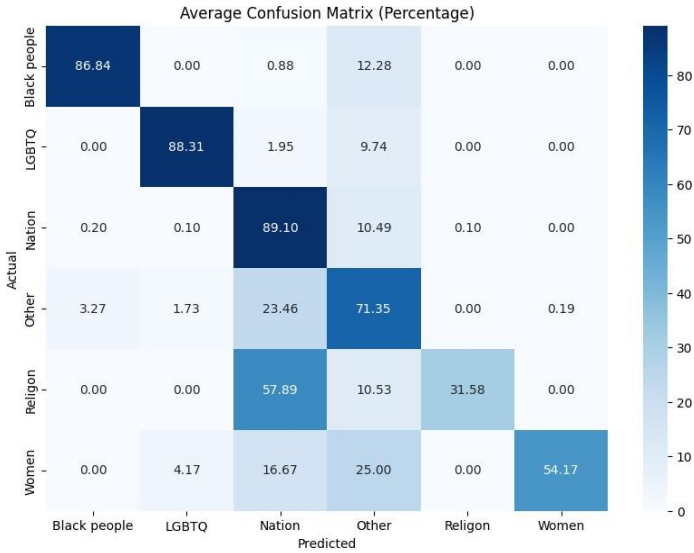

Figure 4: Confusion matrix for the best-performing model in subtask 3 (soft voting + TF-IDF)

- The dataset is relatively small and imbalanced in terms of supportive comments, which has affected the performance of the machine learning models. In future work, this issue could be addressed by increasing the dataset size, particularly for supportive comments. Additionally, techniques should be proposed to manage this imbalanced distribution of data in the experiments.

- The current study reveals that users on YouTube tend to express more support for groups of people than for individuals. Additionally, users show support for different nations without being heavily influenced by religious affiliations. The data indicates that recent support has not been predominantly directed towards any specific religion. Instead, people are more concerned with nations and communities, such as LGBTQ+ individuals and Black people. This trend highlights a broader social focus on national and community identities over religious considerations.

- This study serves as a foundational step in introducing the task of social support detection aimed at fostering support and positivity as an alternative to merely filtering out hate speech Tash et al. (2024a); Ahani et al. (2024b). Consequently, the paper primarily focuses on the introduction of this concept, assessing the feasibility of the task, and examining it from a psychological perspective. As such, experiments involving state-of-the-art models like transformers and large language models have been deferred to future works.

## 10. Conclusion and future work

This study marks a significant step forward in understanding and promoting social support within online communities. By introducing the task of SSD and conducting experiments on a dataset of YouTube comments, we have gained valuable insights into the dynamics of supportive interactions in digital spaces.

Our findings reveal that YouTube users predominantly express support for groups of people, with less emphasis on individual support and religious affiliations. This highlights the importance of considering broader societal contexts when analyzing social support interactions online.

While our experiments have provided promising results, there are notable limitations to address. The dataset used is relatively small and imbalanced, limiting the generalizability of our findings. Additionally, biases inherent in the dataset, stemming from its exclusive focus on YouTube comments within specific parameters, need to be addressed through diversification of data sources.

Looking ahead, future research in SSD should focus on expanding and diversifying datasets, exploring advanced modeling techniques such as deep learning approaches, and designing interventions to promote positive interactions in online communities. By addressing these challenges, we can further our understanding of social support dynamics online and contribute to fostering supportive and inclusive digital spaces.

## Declarations

*Funding*

The work was done with partial support from the Mexican Government through the grant A1-S-47854 of CONACYT,

Mexico, grants 20241816, 20241819, and 20240951 of the Secretaría de Investigación y Posgrado of the Instituto Politécnico Nacional, Mexico. The authors thank the CONACYT for the computing resources brought to them through the Plataforma de Aprendizaje Profundo para Tecnologías del Lenguaje of the Laboratorio de Supercómputo of the INAOE, Mexico and acknowledge the support of Microsoft through the Microsoft Latin America PhD Award.

*Conflict of Interest*

I declare that the authors have no competing interests as defined by Nature Research, or other interests that might be perceived to influence the results and/or discussion reported in this paper.

*Ethics approval*

Not applicable.

*Consent to participate*

Not applicable.

*Consent for publication*

Not applicable.

*Availability of data and materials*

The dataset utilized in this study can be obtained upon request from the corresponding author. Please reach out to Moein Shahiki Tash at [mshahikit2022@cic.ipn.mx].

*Author Contributions Statement*

Z.A. and M.S.T. contributed equally to the work. Z.A. and M.S.T. developed the study concept and design, performed the data analysis, and wrote the main manuscript text. L.R. contributed to data collection, curation, and preprocessing. F.B. provided significant contributions to the methodological framework and data interpretation. G.S. was responsible for software development and technical validation. A.G. supervised the project, provided critical revisions to the manuscript, and contributed to the theoretical framework. All authors reviewed and approved the final manuscript.